\def\year{2021}\relax
\documentclass[letterpaper]{article} 
\usepackage{aaai21}  
\usepackage{times}  
\usepackage{helvet} 
\usepackage{courier}  
\usepackage[hyphens]{url}  
\usepackage{graphicx} 
\usepackage{natbib}  
\usepackage{caption} 
\usepackage{amsmath,amsfonts,bm}

\def\eqref#1{equation~\ref{#1}}

\def\1{\bm{1}}

\def\vh{{\bm{h}}}

\def\mW{{\bm{W}}}
\def\mX{{\bm{X}}}

\DeclareMathAlphabet{\mathsfit}{\encodingdefault}{\sfdefault}{m}{sl}
\SetMathAlphabet{\mathsfit}{bold}{\encodingdefault}{\sfdefault}{bx}{n}

\def\gE{{\mathcal{E}}}

\def\gG{{\mathcal{G}}}
\def\gH{{\mathcal{H}}}

\def\gL{{\mathcal{L}}}

\def\gN{{\mathcal{N}}}

\def\gV{{\mathcal{V}}}

\usepackage{hyperref}
\usepackage{url}
\usepackage{todonotes}
\presetkeys%
    {todonotes}%
    {inline,backgroundcolor=yellow}{}
\usepackage{subcaption}
\usepackage{booktabs}
\usepackage{multirow}
\usepackage{adjustbox}

\title{A Graph-based Imputation Method for Sparse Medical Records}
\author {
    Ramon Viñas,\textsuperscript{\rm 1}
    Xu Zheng, \textsuperscript{\rm 2}
    Jer Hayes \textsuperscript{\rm 2} \\
}
\affiliations {
    \textsuperscript{\rm 1} Department of Computer Science and Technology, University of Cambridge, UK \\
    \textsuperscript{\rm 2} Accenture Labs, Dublin, Ireland \\
    rv340@cam.ac.uk, 
    xu.b.zheng@accenture.com, 
    jeremiah.hayes@accenture.com
}
\begin{document}

\maketitle

\begin{abstract}
Electronic Medical Records (EHR) are extremely sparse. Only a small proportion of events (symptoms, diagnoses, and treatments) are observed in the lifetime of an individual. The high degree of missingness of EHR can be attributed to a large number of factors, including device failure, privacy concerns, or other unexpected reasons. Unfortunately, many traditional imputation methods are not well suited for highly sparse data and scale poorly to high dimensional datasets. In this paper, we propose a graph-based imputation method that is both robust to sparsity and to unreliable unmeasured events. Our approach compares favourably to several standard and state-of-the-art imputation methods in terms of performance and runtime. Moreover, results indicate that the model learns to embed different event types in a clinically meaningful way. Our work can facilitate the diagnosis of novel diseases based on the clinical history of past events, with the potential to increase our understanding of the landscape of comorbidities.

\end{abstract}

\section{Introduction}\label{intro}
Missing data is a pervasive challenge for medical domains that can result in the reduction of the statistical power of a study and can produce biased estimates, leading to invalid conclusions \citep{Hyun2013}. Previous research has demonstrated success in data imputation with both statistical and generative models based approaches. These traditional data imputation methods include univariate methods such as mean or median imputation, multi-variate methods like k-NN imputation \citep{troyanskaya2001missing} and Multivariate Imputation by Chained Equations (MICE) \citep{van2011mice}, and deep learning methods such as autoencoders \citep{vincent2008extracting}, GAIN \citep{yoon2018gain}, and GRAPE \citep{grape}.

Although these methods show good performance on traditional datasets, they do not explicitly deal with the sparsity and imbalance problem that characterises many medical datasets. From a machine learning perspective, medical data can potentially be very sparse, e.g. there are over 69,000 International Classification of Diseases (ICD) diagnosis codes and even a very diseased patient will only present a minuscule fraction of these codes. Together with high dimensionality, the sparsity issue often  makes it infeasible to apply traditional data imputation methods on medical datasets. 

In recent years, Graph Neural Networks (GNN) have gained increasing attention for modelling graph-based datasets, including social networks, citation networks, and molecules. In this paper, we build on GRAPE \citep{grape} to develop a GNN-based imputation method for highly imbalanced medical datasets. Essentially, GRAPE represents data as a bipartite graph where sample and feature nodes are connected according to the missingness pattern of the dataset. Data imputation then corresponds to edge-level prediction for the missing links. To deal with the highly sparse and unary nature of EHRs (i.e. missing values represent either unmeasured events or negative outcomes), we employ a training scheme that balances positive and unmeasured edges in the graph. We perform experiments on EHRs from a subset of more than 72,000 diabetes patients from the IBM Explorys dataset. Our results show improved performance with respect to traditional and state-of-the-art methods that do not explicitly deal with the imbalanced and unary nature of the data. Besides, we demonstrate that the latent embeddings of medical concepts retrieved from our model cluster in a clinically meaningful way.


\section{Method}\label{method}
\paragraph{Problem definition. } Let $\mX \in \{0, 1\}^{m \times n}$ be a medical dataset with $m$ patients and $n$ event types (e.g. diagnoses). For each patient $i$ and event type $j$, the entry $x_{ij}$ is one if event $i$ (e.g. familial Mediterranean fever) has been observed for patient $j$ and zero otherwise (i.e. unmeasured events with unknown outcomes). Here we assume that the entries $x_{ij}$ are binary (unmeasured and measured), but our approach can be extended to account for more complex variable types (e.g. sequence of event times). Our goal is to impute the unmeasured values of dataset $\mX$.

\paragraph{Bipartite graph representation. } Following \citep{grape}, we represent data in a bipartite graph $\gG = \{\gV_p \cup \gV_e, \gE\}$, where $\gV_p = \{v^p_1, ..., v^p_m\}$ is the patient partition, $\gV_e = \{v^e_1, ..., v^e_n\}$ is the event partition, and $\gE = \{(v^p_i, v^e_j) | v^p_i \in \gV_p, v^e_j \in \gV_e, x_{ij} = 1 \}$ is the set of edges connecting patients from $\gV_p$ to events from $\gV_e$ according to the measured entries of the dataset. This framework also allows attributed edges (e.g. with event times) with potential edge repetitions (i.e. a certain event can occur several times for the same patient). 

\paragraph{Model. } We employ a graph neural network to perform link prediction on the bipartite graph. This procedure can be divided into 3 steps.

First, we initialise node features of the patient $\gV_p$ and event nodes $\gV_e$. For event nodes, we use $d$-dimensional learnable embeddings as initial node values. The idea is that these weights, which will be learnt through gradient descent, should summarise relevant properties of each event. For patient nodes, we initialise the node features with the available demographic information (e.g. age and sex) and project them to the $d$-dimensional space with a multi-layer perceptron. Importantly, this formulation allows transfer learning between sets of distinct patients.

Second, we perform message passing to compute latent node embeddings. Let $\gH_p = \{\vh^p_{1}, ..., \vh^p_{m} \}$ and $\gH_e = \{\vh^e_{1}, .., \vh^e_{n} \}$ be the initial $k$-dimensional patient and event node embeddings, respectively. Let $\gN_p(i; \gE) = \{j | (v^p_i, v^e_j) \in \gE\} $ and $\gN_e(i; \gE) = \{j | (v^p_j, v^e_i) \in \gE\}$ be the set of neighbours of nodes $v^p_i$ and $v^e_i$, respectively. We compute latent node embeddings $\hat{\gH}_p = \{\hat{\vh}^p_{1}, ..., \hat{\vh}^p_{m}\}$ and $\hat{\gH}_e = \{\hat{\vh}^e_{1}, ..., \hat{\vh}^e_{n}\}$ with separate GraphSAGE layers \citep{hamilton2018inductive} as follows:
\begin{equation}
    \begin{split}
        \hat{\vh}^p_i = \mW^p_1 \vh^p_i + \mW^p_2 \Bigg(\frac{1}{|\gN_p(i; \gE)|} \sum_{j \in \gN_p(i)} \vh^e_j \Bigg) \\
        \hat{\vh}^e_i = \mW^e_1 \vh^e_i + \mW^e_2 \Bigg(\frac{1}{|\gN_e(i, \gE)|} \sum_{j \in \gN_e(i)} \vh^p_j \Bigg)
    \end{split}
\end{equation}
where $\mW^p_1$, $\mW^p_2$, $\mW^e_1$, $\mW^e_2$ are learnable weights. Optionally, we can stack several layers interleaving non-linearities.


Finally, unmeasured values are imputed via link prediction. We compute the probability of an edge between nodes $v^p_i$ and $v^e_j$ as $p(x_{ij} = 1) = \text{MLP}(\hat{\vh}^p_i, \hat{\vh}^e_j)$, where $\text{MLP}$ is a multi-layer perceptron with a sigmoid function as output activation.

\begin{figure}
    \centering
    \includegraphics[width=.45\textwidth, trim={0.2cm 0 0 0},clip]{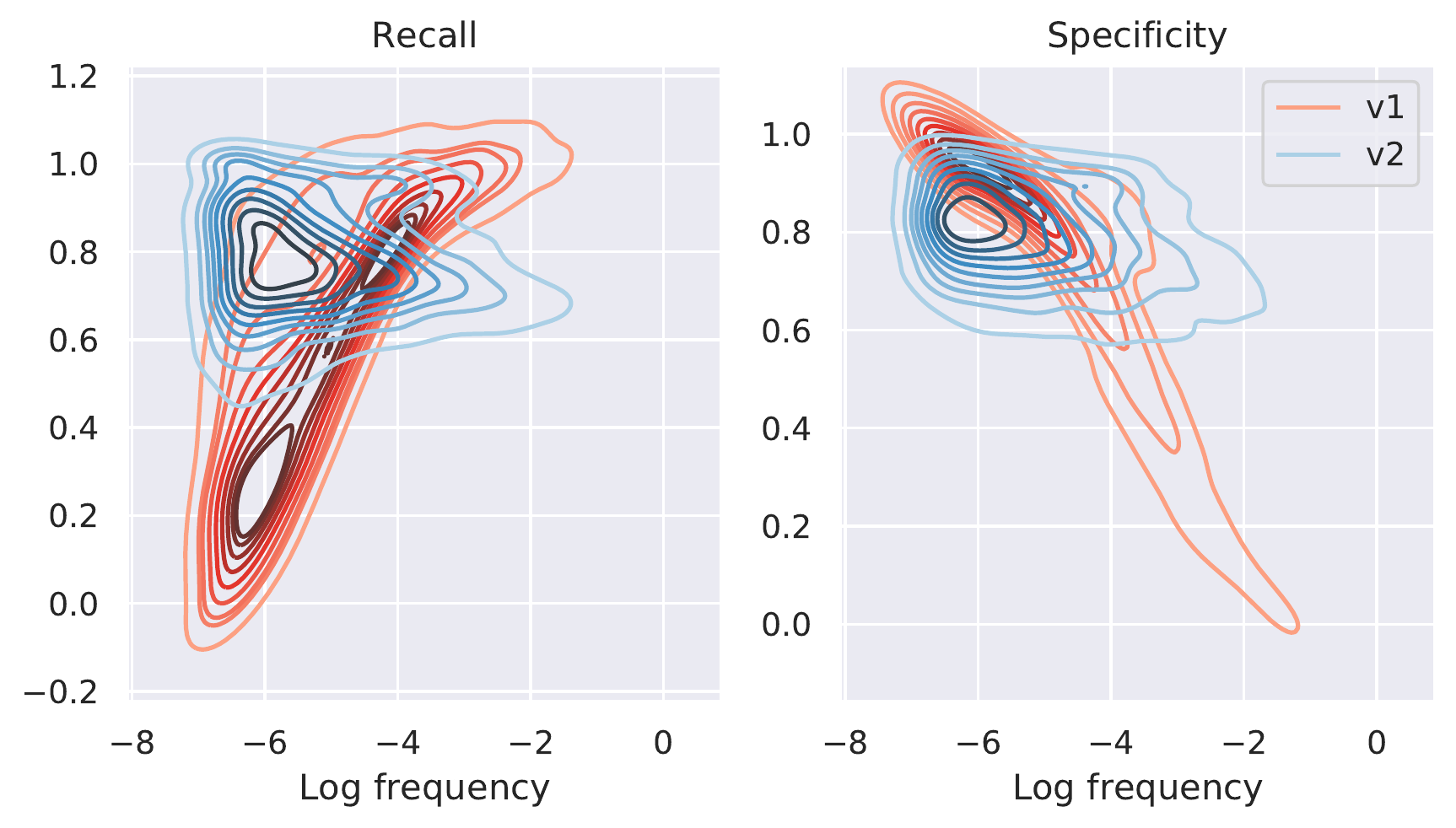}
    \caption{Distribution of per-event recalls and specificities (y-axis) by event frequencies (x-axis),  computed from the graph-based model trained under two undersampling schemes. Version 1 (red): the number of events in the invisible and negative sets match (i.e. $|\gE_{\text{inv}}| = |\gE_{\text{neg}}|$). Version 2 (blue, ours): the event and patient frequencies are additionally preserved (i.e. $| \gN_p(i; \gE_{\text{neg}}) | = | \gN_p(i; \gE_{\text{inv}}) | $ and $| \gN_e(j; \gE_{\text{neg}}) | = | \gN_e(j; \gE_{\text{inv}}) | $ for any patient $i$ and event type $j$). Version 1 (red) introduces a sampling bias -- common events are rarely imputed as negative, while rare events are rarely imputed as active.}
    \label{fig:sampling_bias}
\end{figure}


\paragraph{Optimisation. } To deal with the highly unbalanced data, we employ a training scheme that balances positive and unmeasured edges. At each training iteration, we sample three sets of edges:
\begin{itemize}
    \item \textbf{Invisible edges}. The set $\gE_{\text{inv}}$ contains $k$ positive edges from $\gE$ that are unseen to the model, where $k$ is a hyperparameter. The goal is to correctly predict the presence of these edges via link prediction.
    \item \textbf{Visible edges}. The set $\gE_{\text{vis}}$ contains the remaining $|\gE| - k$ positive edges that are seen during message passing. The visible and invisible subsets are disjoint and $\gE = \gE_{\text{vis}} \cup \gE_{\text{inv}}$.
    \item \textbf{Negative edges}. The set $\gE_{\text{neg}}$ contains $k$ edges that do not belong to $\gE$. The goal of the model is to correctly predict the absence of these edges. Importantly, the cardinalities of $\gE_{\text{neg}}$ and $\gE_{\text{inv}}$ match. Moreover, the set $\gE_{\text{neg}}$ preserves the patients and event frequencies, i.e. $| \gN_p(i; \gE_{\text{neg}}) | = | \gN_p(i; \gE_{\text{inv}}) | $ and $| \gN_e(j; \gE_{\text{neg}}) | = | \gN_e(j; \gE_{\text{inv}}) | $ for any patient $i$ and event type $j$. This effectively balances the model’s exposure to positive and unmeasured edges for each patient and event type, preventing any sampling biases (see Figure \ref{fig:sampling_bias}).
\end{itemize}

We then optimise model's parameters via gradient descent by minimising the binary cross-entropy:
\begin{equation}
    \begin{split}
    \gL(\gE_{\text{inv}}, \gE_{\text{neg}}) =& -\frac{1}{2k}\sum_{(v^p_i, v^e_j) \in \gE_{\text{inv}}} \log p(x_{ij} = 1) \\
    & - \frac{1}{2k} \sum_{(v^p_i, v^e_j) \in \gE_{\text{neg}}} \log p(x_{ij} = 0)
    \end{split}
\end{equation}

\section{Results}
\label{exp}

\paragraph{Evaluation metrics. } Medical records are extremely sparse -- only a very small proportion of events (e.g. symptoms, diagnoses, treatments) are observed during the lifetime of an individual. At the same time, we cannot always be certain that unobserved events have not occurred because in practise we can only measure a small fraction of them (e.g., it is unfeasible to test someone for all known diseases) -- this is precisely why we want to impute missing values. Conversely, observed events have happened in reality with high confidence (e.g. chemotherapy for lung cancer). In this paper, we evaluate the imputation performance with sensitivity, specificity, and balanced accuracy. In contrast to accuracy (uninformative in sparse scenarios) and precision (sensitive to unreliable false positives), the proposed metrics are both robust to sparsity and to unreliable unmeasured events.

\paragraph{Dataset. } Patients with diabetes are sampled from the the IBM Explorys database. We create a bipartite graph of patients and events using diagnoses-related events. We filter out events that appear in less than $0.1\%$ of the records, resulting in 3284 unique events. We split patients into disjoint train ($70\%$) and test ($30\%$) sets, yielding 72801 and 30334 unique train and test patients, respectively. For the test patients, we mask out $30\%$ of the observed values and use them to evaluate the performance of all the models. We leverage the age and sex of the patients as demographic information provided as input to the models. When represented as a matrix, the dataset is highly sparse, with 98.3\% of zero entries. 

\paragraph{Baseline models. } We compare our model to several baseline methods, including k-NN imputation \citep{troyanskaya2001missing}, Generative Adversarial Imputation Networks (GAIN) \cite{yoon2018gain} and Denoising Autoencoders (DAE) \cite{vincent2008extracting}. As these baseline models can only handle tabular data, we represent patient records as a binary matrix where rows correspond to patients and columns to unique diagnosis codes. In this matrix, entry $(i, j)$ is one if the $j$-th diagnosis has been observed for patient $i$-th and zero otherwise. 


The denoising autoencoder (DAE) and generative adversarial imputation networks (GAIN) are both optimised via the reconstruction error on the observed values (plus an adversarial term for the missing values for GAIN). Because the dataset is highly imbalanced, both models are by default biased towards the majority class, i.e. zero for each feature. Additionally, GAIN cannot readily deal with the unary nature of the data -- missing positive values cannot be distinguished from actual zeros (i.e. events with negative outcome) and they both form the mask vector. To address these issues, we adopt an undersampling mechanism that closely mimicks the training scheme of the GNN model. At each training iteration, we randomly sample $k$ negative values (i.e. unmeasured events), where $k$ is the total number of positive values, and treat them as observable. The remaining entries are masked out (and form the mask vector for GAIN) and the methods are optimised by minimising their respective loss functions.

\paragraph{Hyperparameters. } We use node embeddings of dimension $d=95$. We initialise them with the right singular vectors of the train dataset computed via singular value decomposition (SVD). This yields higher validation scores according to our experiments. The graph neural network architecture consists of 3 GraphSAGE layers with node embeddings of dimension $d=95$ and rectified linear unit (ReLU) activations. The final multi-layer perceptron comprises 1 hidden layer with 32 units followed by ReLU. We optimise the model with the Adam optimiser \citep{kingma2014adam} and a learning rate of $0.0066$. At each training iteration, we randomly sample the invisible $\gE_{\text{inv}}$ set of edges from a binomial distribution $B(1, p)$ with probability $p=0.2$, that is, on average $20\%$ of the total number of training edges are masked out. The set $\gE_{\text{neg}}$ of unmeasured edges is then sampled as described in the optimisation section, preserving the cardinality of $\gE_{\text{inv}}$. We implement the model in Pytorch \citep{pytorch} and Pytorch Geometric \citep{Fey/Lenssen/2019}.

\begin{figure}
    \centering
    \includegraphics[width=.5\textwidth, trim={2.5cm 2.2cm 1cm 2.2cm},clip]{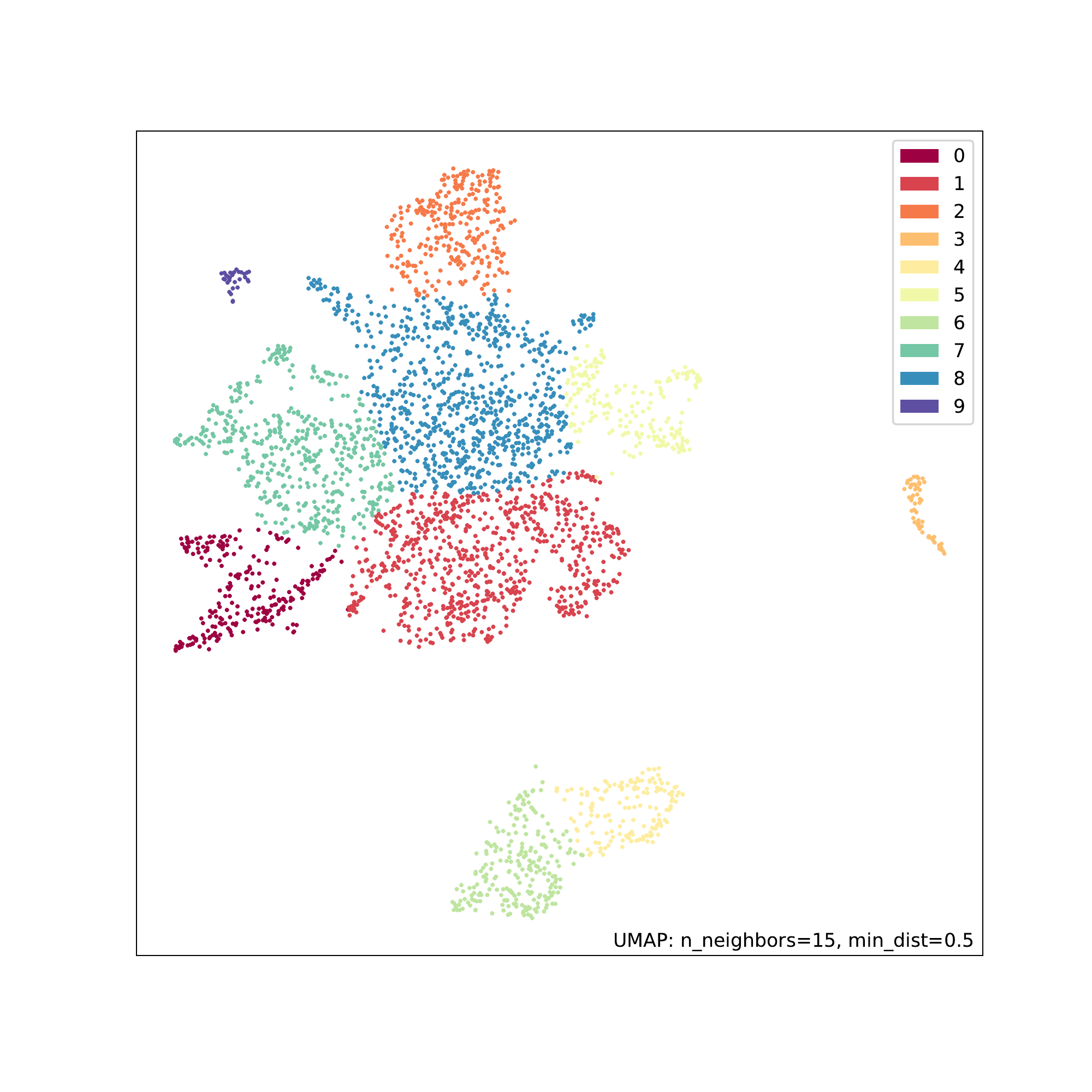}
    \caption{Low-dimensional representation of the latent event embeddings. Each point corresponds to a different event type (e.g. malnutrition-related diabetes mellitus). We use spectral clustering to cluster events into 10 clusters and analyse their composition in Figure \ref{fig:cluster_composition}. Overall, we find that the model is able to identify clusters of clinically-related events. }
    \label{fig:UMAP}
\end{figure}

\begin{figure*}[ht]
    \centering
    \includegraphics[width=\textwidth]{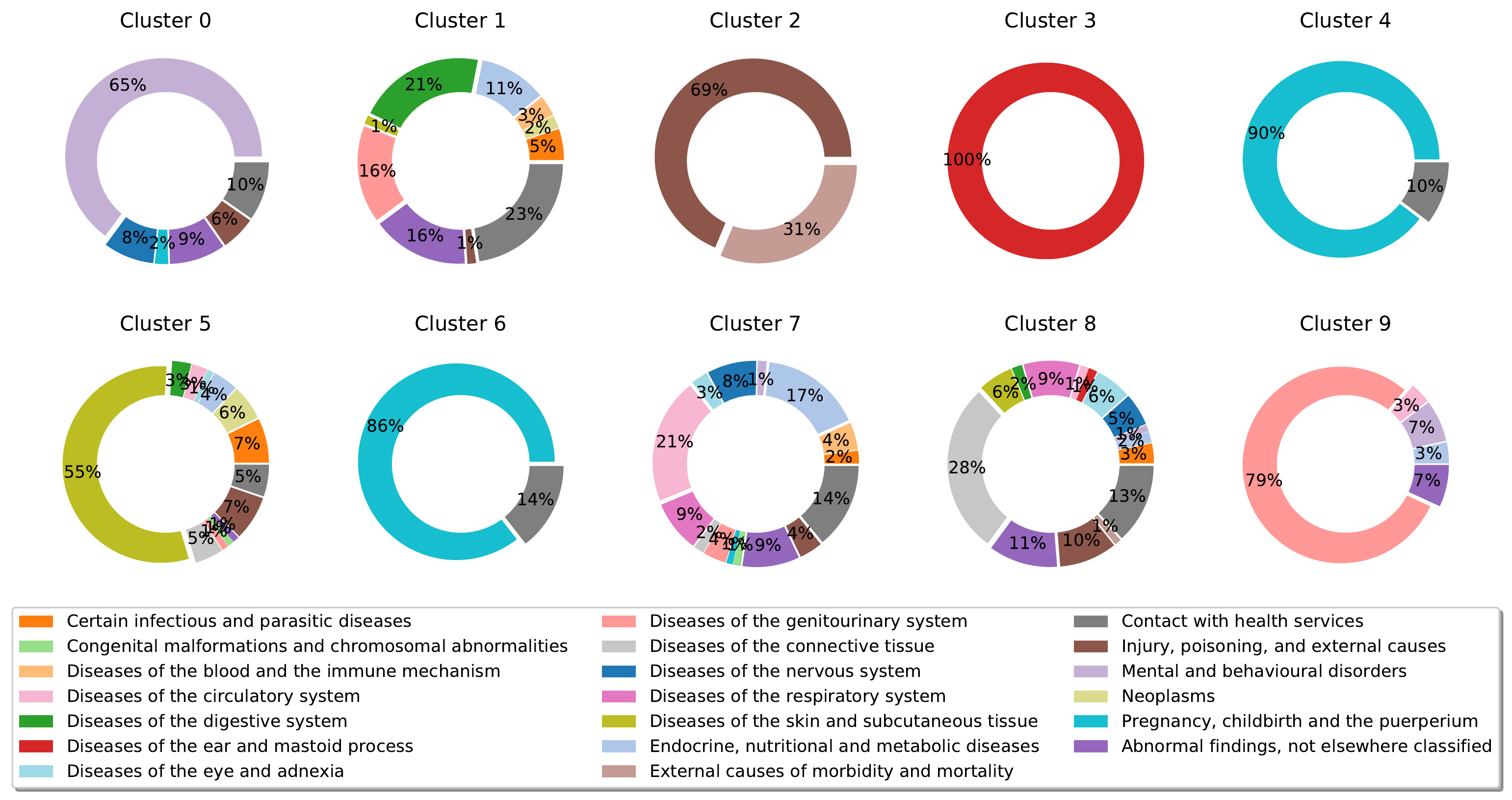}
    \caption{Composition of low-dimensional event clusters. We inspect the event-type composition of clusters identified via spectral clustering on the latent event embeddings of our imputation model (see Figure \ref{fig:UMAP}). We observe high cluster purity, that is, several clusters consist mostly of events of the same type. For example, 65\% of events in cluster 0 are mental and behavioural related disorders, while most events in clusters 4 and 6 are related to pregnancy and childbirth. Other clusters are more heterogeneous and understanding their semantics would possibly require a finer-grained analysis. }
    \label{fig:cluster_composition}
\end{figure*}

\paragraph{Results. } Table \ref{tab:results} shows the test imputation scores. We compute the sensitivity, specificity, and balanced accuracy for each event type and report the mean standard deviation of these metrics. Importantly, conventional imputation methods such as k-NN imputation do not have any built-in mechanisms to deal with the inherent characteristics of medical records (i.e. sparsity and unreliability of unmeasured events) and model calibration is therefore unclear. For these methods, we analyse their performance under two different thresholds, namely a $0.5$ cutoff (i.e. for patient $i$, event $j$ is imputed as measured if $p(x_{ij} = 1) > 0.5$) and the per-event-type frequencies of measured values in the train set (i.e. for patient $i$, event $j$ is imputed as measured if $p(x_{ij} = 1) > \frac{1}{m} \sum_{i=1}^m x_{ij}$, where $m$ is the number of train patients). Despite the data sparsity, the GNN-based method attains highly balanced predictions (sensitivity$=0.78\pm0.12$, specificity$=0.79\pm0.09$) with the default $0.5$ cutoff, outperforming other baselines by a large margin in terms of balanced accuracy (arithmetic mean of sensitivity and specificity). We attribute this to the training scheme, which effectively balances the model's exposure to positive and unmeasured edges for each event type, yielding a well-calibrated model (see Figure \ref{fig:sampling_bias} for a comparison with a random undersampling method). The proposed approach is also highly scalable and significantly faster than traditional methods.


\begin{table}[ht]
    \centering
    \caption{Imputation results. We report the mean and standard deviation of the per-event-type evaluation metric for different cutoff values. For the threshold \emph{Avg.}, we use the per-event-type frequencies of observed values in the train set as cutoffs. The best scores are highlighted in bold.}
    \label{tab:results}
    \resizebox{\columnwidth}{!}{%
    \begin{tabular}{cccccc}
        \toprule
        Method & Cutoff & Sensitivity & Specificity & Balanced Acc. & Runtime \\
        \midrule
        \multirow{2}{*}{10-NN} & $0.5$ & $0.08 \pm 0.20$ & $\mathbf{0.98 \pm 0.10}$ & $0.53 \pm 0.07$ & \multirow{2}{*}{$43.29$h}\\
         & Avg. & $\underline{0.63 \pm 0.29}$ & $0.63 \pm 0.26$ & $\underline{0.63 \pm 0.09}$\\
        \multirow{2}{*}{DAE} & $0.5$ & $0.36 \pm 0.34$ & $\underline{0.87 \pm 0.24}$  & $0.62 \pm 0.11$ & \multirow{2}{*}{$0.25$h} \\
        & Avg. & $\mathbf{0.98\pm 0.08}$ & $0.09 \pm 0.18$  & $0.54 \pm 0.08$  \\
        \multirow{2}{*}{GAIN} & $0.5$ & $0.36 \pm 0.29$ & $0.83 \pm 0.25$  & $0.59 \pm 0.11$ & \multirow{2}{*}{$0.25$h} \\
        & Avg. & $0.45\pm 0.31$ & $0.73 \pm 0.29$  & $0.59 \pm 0.11$ \\
        \midrule
        Ours & $0.5$ & $0.78 \pm 0.12$ & $0.79 \pm 0.09$ & $\mathbf{0.79 \pm 0.09}$ & $1.09$h \\
        \bottomrule
    \end{tabular}%
    }
\end{table}


We further study the behaviour of our model by inspecting the latent event embeddings obtained after message passing. We employ UMAP \cite{mcinnes2018umap} to project these embeddings into a 2-dimensional space and further apply spectral clustering ($N$=10 clusters) to cluster events (see Figure \ref{fig:UMAP}). Figure \ref{fig:cluster_composition} depicts their event type composition. Cluster 0 mostly consists of mental and behavioural related disorders (65\%). Events in cluster 2 involve injuries and poisoning (69\%) and external causes of morbidity and mortality (31\%). Cluster 3 is composed exclusively by diseases of the ear and mastoid process. Clusters 4 and 6 are both related to pregnancy, childbirth, and the puerperium. The majority of events in cluster 5 are diseases of the skin and the subcutaneous tissue (55\%), while most events in cluster 9 are diseases of the genitourinary system. The remaining clusters are more heterogeneous and understanding their semantics would possibly require a finer-grained analysis (e.g. cluster 1 consists of events related to contact with health services, 23\%, and diseases of the digestive system, 21\%, among others). Overall, this analysis shows that the model is grouping clinically-related event types in the latent space, with high cluster purity. 


\section{Conclusion}

In this paper, we have studied the problem of imputing missing data in medical records. These datasets are highly sparse and unmeasured events are unreliable (i.e. the fact that a specific event has not been observed for a certain patient does not entail that it has not occurred in reality). Unfortunately, traditional imputation methods are not well suited for this scenario. To address this challenge, we have proposed a graph-based deep learning model that is both scalable and effective at imputing missing values in sparse regimes. The proposed model is easy to use and well-calibrated by default. Furthermore, our approach compares favourably to existing methods in terms of performance and runtime. This work can facilitate the diagnosis of new events and shed light into the landscape of comorbidities.

\bibliography{bibliography}

\begin{thebibliography}{11}
\providecommand{\natexlab}[1]{#1}
\providecommand{\url}[1]{\texttt{#1}}
\providecommand{\urlprefix}{URL }
\expandafter\ifx\csname urlstyle\endcsname\relax
  \providecommand{\doi}[1]{doi:\discretionary{}{}{}#1}\else
  \providecommand{\doi}{doi:\discretionary{}{}{}\begingroup
  \urlstyle{rm}\Url}\fi

\bibitem[{Fey and Lenssen(2019)}]{Fey/Lenssen/2019}
Fey, M.; and Lenssen, J.~E. 2019.
\newblock Fast Graph Representation Learning with {PyTorch Geometric}.
\newblock In \emph{ICLR Workshop on Representation Learning on Graphs and
  Manifolds}.

\bibitem[{Hamilton, Ying, and Leskovec(2018)}]{hamilton2018inductive}
Hamilton, W.~L.; Ying, R.; and Leskovec, J. 2018.
\newblock Inductive Representation Learning on Large Graphs.

\bibitem[{Hyun(2013)}]{Hyun2013}
Hyun, K. 2013.
\newblock The prevention and handling of the missing data.
\newblock \emph{Korean J Anesthesiol} 64(5): 402--406.

\bibitem[{Kingma and Ba(2014)}]{kingma2014adam}
Kingma, D.~P.; and Ba, J. 2014.
\newblock Adam: A method for stochastic optimization.
\newblock \emph{arXiv preprint arXiv:1412.6980} .

\bibitem[{McInnes, Healy, and Melville(2018)}]{mcinnes2018umap}
McInnes, L.; Healy, J.; and Melville, J. 2018.
\newblock Umap: Uniform manifold approximation and projection for dimension
  reduction.
\newblock \emph{arXiv preprint arXiv:1802.03426} .

\bibitem[{Paszke et~al.(2019)Paszke, Gross, Massa, Lerer, Bradbury, Chanan,
  Killeen, Lin, Gimelshein, Antiga, Desmaison, Kopf, Yang, DeVito, Raison,
  Tejani, Chilamkurthy, Steiner, Fang, Bai, and Chintala}]{pytorch}
Paszke, A.; Gross, S.; Massa, F.; Lerer, A.; Bradbury, J.; Chanan, G.; Killeen,
  T.; Lin, Z.; Gimelshein, N.; Antiga, L.; Desmaison, A.; Kopf, A.; Yang, E.;
  DeVito, Z.; Raison, M.; Tejani, A.; Chilamkurthy, S.; Steiner, B.; Fang, L.;
  Bai, J.; and Chintala, S. 2019.
\newblock PyTorch: An Imperative Style, High-Performance Deep Learning Library.
\newblock In Wallach, H.; Larochelle, H.; Beygelzimer, A.; d\textquotesingle
  Alch\'{e}-Buc, F.; Fox, E.; and Garnett, R., eds., \emph{Advances in Neural
  Information Processing Systems 32}, 8024--8035. Curran Associates, Inc.
\newblock
  \urlprefix\url{http://papers.neurips.cc/paper/9015-pytorch-an-imperative-style-high-performance-deep-learning-library.pdf}.

\bibitem[{Troyanskaya et~al.(2001)Troyanskaya, Cantor, Sherlock, Brown, Hastie,
  Tibshirani, Botstein, and Altman}]{troyanskaya2001missing}
Troyanskaya, O.; Cantor, M.; Sherlock, G.; Brown, P.; Hastie, T.; Tibshirani,
  R.; Botstein, D.; and Altman, R.~B. 2001.
\newblock Missing value estimation methods for DNA microarrays.
\newblock \emph{Bioinformatics} 17(6): 520--525.

\bibitem[{Van~Buuren and Groothuis-Oudshoorn(2011)}]{van2011mice}
Van~Buuren, S.; and Groothuis-Oudshoorn, K. 2011.
\newblock mice: Multivariate imputation by chained equations in R.
\newblock \emph{Journal of statistical software} 45: 1--67.

\bibitem[{Vincent et~al.(2008)Vincent, Larochelle, Bengio, and
  Manzagol}]{vincent2008extracting}
Vincent, P.; Larochelle, H.; Bengio, Y.; and Manzagol, P.-A. 2008.
\newblock Extracting and composing robust features with denoising autoencoders.
\newblock In \emph{Proceedings of the 25th international conference on Machine
  learning}, 1096--1103.

\bibitem[{Yoon, Jordon, and Schaar(2018)}]{yoon2018gain}
Yoon, J.; Jordon, J.; and Schaar, M. 2018.
\newblock Gain: Missing data imputation using generative adversarial nets.
\newblock In \emph{International Conference on Machine Learning}, 5689--5698.
  PMLR.

\bibitem[{You et~al.(2020)You, Ma, Ding, Kochenderfer, and Leskovec}]{grape}
You, J.; Ma, X.; Ding, D.; Kochenderfer, M.; and Leskovec, J. 2020.
\newblock Handling Missing Data with Graph Representation Learning.
\newblock \emph{NeurIPS} .

\end{thebibliography}


\end{document}